\documentclass{article}
\usepackage{xcolor}

\PassOptionsToPackage{numbers}{natbib}
\usepackage[dblblindworkshop, final]{neurips_2025}  

\usepackage[utf8]{inputenc} 
\usepackage[T1]{fontenc}    
\usepackage{color}         
\usepackage{hyperref}       
\usepackage{colortbl}      
\usepackage{adjustbox}     
\usepackage{textcomp}      
\usepackage{amsmath}
\usepackage{float}          
\usepackage{url}            
\usepackage{booktabs}       
\usepackage{multirow}
\usepackage{amsfonts}
\usepackage[table]{xcolor}
\usepackage{nicefrac}       
\usepackage{microtype}      
\usepackage{todonotes}

\usepackage{amssymb} 

\newcommand{\databg}[1]{\cellcolor{green!20}{#1}}

\title{Adaptive Parameter Optimization for Robust Remote Photoplethysmography}
\workshoptitle{Learning from Time Series for Health}

\author{%
\begin{tabular}{ccc}
 Cecilia G. Morales\textsuperscript{1*$\dagger$} & Fanurs Chi-En Teh\textsuperscript{1*$\dagger$} & Kai Li\textsuperscript{1,2*} \\
 \texttt{cgmorale@andrew.cmu.edu} & \texttt{fanurs@cmu.edu} &
 \texttt{kaii.li@mail.utoronto.ca}\\
\end{tabular}
\\[3ex] 
\begin{tabular}{cc}
\textbf{Pushpak Agrawal}\textsuperscript{1,3} & \textbf{Artur Dubrawski}\textsuperscript{1} \\
\texttt{pushpaka@andrew.cmu.edu} & \texttt{awd@cs.cmu.edu} \\
\end{tabular}
\\[3ex]
\textsuperscript{1}Carnegie Mellon University, USA \quad
\textsuperscript{2}University of Toronto, Canada \quad
\textsuperscript{3}Vellore Institute of Technology
}

\begin{document}
\maketitle

\renewcommand{\thefootnote}{\fnsymbol{footnote}}
\footnotetext[1]{Co-first authors contributed equally to this work.}
\footnotetext[2]{Corresponding authors}
\renewcommand{\thefootnote}{\arabic{footnote}}

\begin{abstract}
Remote photoplethysmography (rPPG) enables contactless vital sign monitoring using standard RGB cameras. However, existing methods rely on fixed parameters optimized for particular lighting conditions and camera setups, limiting adaptability to diverse deployment environments. This paper introduces the Projection-based Robust Signal Mixing (PRISM) algorithm, a training-free method that jointly optimizes photometric detrending and color mixing through online parameter adaptation based on signal quality assessment. PRISM achieves state-of-the-art performance among unsupervised methods, with MAE of 0.77 bpm on PURE and 0.66 bpm on UBFC-rPPG, and accuracy of  97.3\% and 97.5\% respectively at a 5 bpm threshold. Statistical analysis confirms PRISM performs equivalently to leading supervised methods ($p > 0.2$), while maintaining real-time CPU performance without training. This validates that adaptive time series optimization significantly improves rPPG across diverse conditions.
\end{abstract}



\section{Introduction}
\label{sec:introduction}

Remote photoplethysmography (rPPG) enables contactless cardiac monitoring by detecting pulse-induced color variations on skin using standard RGB cameras to extract physiological time series~\cite{10.1145/3558518}. At its core, rPPG requires combining raw RGB channels into a single pulse signal, with the combination strategy directly affecting performance and generalizability across applications such as telemedicine consultations and spotting driver drowsiness~\cite{lee2024review}. While supervised deep models can adapt to diverse conditions, they require extensive domain-specific datasets and substantial computation~\cite{ni2021review}. Current unsupervised methods remain interpretable and computationally efficient but rely on fixed parameters, limiting their adaptability to real-world lighting variations~\cite{DiLernia2024rPPG}.


The plane-orthogonal-to-skin (POS) algorithm exemplifies this trade-off: it removes low-frequency drift, suppresses specularities via a projection orthogonal to mean skin color, and mixes channels under an assumption of stable spectral composition with varying intensity~\cite{pos}. Such assumptions hold reasonably well indoors under artificial lighting (fluorescent/incandescent), but natural illumination is nonstationary; the incident spectrum varies with solar angle, atmosphere, shadows, and scene context, so fixed projection axes can fail when spectra and luminance covary~\cite{8037807}. Invariance-based approaches such as LGI mitigate some nuisance variation but remain static once chosen~\cite{lgi}.

We build on the POS algorithm’s structure by retaining its interpretable projection design, but replace the fixed color mixing with two adaptive parameters, $\alpha$ and $\lambda$, which are optimized per window to maximize heart rate signal quality. This principled objective yields a globally pulse-dominated yet temporally smooth signal that resists illumination shifts and motion, while remaining fully interpretable and CPU-efficient. It preserves POS's strengths while significantly boosting accuracy for autonomous or human-in-the-loop care.

This paper introduces PRISM, a training-free method that adaptively optimizes rPPG signal extraction through online parameter selection. We demonstrate state-of-the-art unsupervised performance in PURE and comparable to supervised methods in UBFC-rPPG while maintaining real-time efficiency.

\section{Related Works}
While supervised deep learning approaches have shown promising results in rPPG, they face significant challenges in generalization across diverse populations, lighting conditions, and camera systems due to their dependence on large labeled datasets~\cite{cheng2021deep}. Unsupervised methods address these limitations by relying on fundamental physiological and optical principles rather than learned patterns. Since PRISM is a training-free algorithm, we focus this review on unsupervised rPPG approaches that are computationally efficient enough for real-time processing ~\cite{lee2024review}. 
These methods typically rely on color space transformations, blind source separation, and physiological priors to extract the pulse signal from raw video data~\cite{debnath2025comprehensive}.

De Haan and Jeanne proposed the CHROM method, which used a skin reflection model to compute orthogonal chrominance signals, achieving 92\% agreement with contact PPG and improved motion robustness over blind source separation like ICA and PCA~\cite{chrom}. Wang et al. extended this with 
the Plane-Orthogonal-to-Skin (POS) method, leveraging optical and physiological skin modeling. POS outperformed CHROM, Green, PCA, and ICA in varying skin tones and activity levels in controlled settings~\cite{pos}.

More recent methods include Orthogonal Matrix Image Transformation (OMIT) using QR decomposition and Local Group Invariance (LGI) with signal priors. LGI was designed to handle adaptive lighting via invariance priors~\cite{lgi}, it still relies on predetermined algorithmic choices that may not generalize across all conditions. PRISM addresses these limitations through a different approach. Rather than using fixed parameters or invariance assumptions,
PRISM maintains POS's interpretable green minus red/blue projection while enabling dynamic optimization rather than fixed parameter selection. 
Unlike existing approaches, PRISM provides a unified framework that continuously adapts both color mixing and temporal filtering parameters based on real-time signal quality, ensuring optimal performance across diverse lighting conditions, motion artifacts, and individual physiological variations without requiring parameter tuning. 

\section{Methods}
\label{sec:methods}

PRISM adaptively selects projection and detrending parameters based on a signal quality objective. It extends the structure of the Plane-Orthogonal-to-Skin (POS) algorithm~\cite{pos}, while replacing its fixed projection and filtering parameters with ones optimized online for each video.

Time series of RGB values are extracted by averaging over a facial region of interest detected and tracked throughout the video using YOLOv5~\cite{yolov5}. Each raw channel $X(t) \in \{R(t), G(t), B(t)\}$ is normalized by a smooth spline baseline $\tilde X(t)$, yielding $\hat X(t) = X(t)/\tilde X(t)$, to suppress slow changes from illumination and sensor gain while preserving fine-grained pulse signals. A smoothness parameter $\lambda > 0$ controls the spline, expressed in units of $\text{s}^3$ throughout this text.

To extract the pulse signal, we form a linear combination of the normalized color channels. The POS method projects RGB signals into a chrominance plane orthogonal to the intensity direction and combines two empirically chosen components using adaptive weighting. PRISM preserves this structure but restricts to a single axis within that plane, forming a direct projection:
\begin{equation}
    s(t) = \hat G(t) - \left( \alpha \hat B(t) + (1 - \alpha) \hat R(t) \right) \ ,
\end{equation}
where $\alpha \in [0, 1]$ controls the red–blue balance. This construction remains interpretable and allows adaptive selection of the projection axis. The pulse signal $s(t)$ is then split into non-overlapping windows of $\delta t = 10$ seconds, and we extract a heart rate estimate $h_i$ from each.

To select optimal parameters $(\lambda^*, \alpha^*)$, we define an objective that balances two properties of the signal $s(t)$: spectral concentration within the physiological band and temporal stability of the estimated heart rate. Specifically, we minimize:
\begin{equation}
    (\lambda^*, \alpha^*) = \arg\min_{\lambda, \alpha} \left[ k \cdot \mathrm{TV}(\lambda, \alpha) - C(\lambda, \alpha) \right] \ ,
\end{equation}
where $k = 1/3$ is set empirically, $C$ measures the fraction of power concentrated in the plausible heart rate band (typically 0.5-4.0 Hz), and $\mathrm{TV}$ is the temporal variation that quantifies how much the heart rate estimates vary between consecutive windows, defined as $\mathrm{TV} = \frac{1}{t_n - t_1} \sum_{i=1}^{n - 1} | h_{i+1} - h_i | $.

This encourages both sharp spectral peaks and smoothly evolving heart rate sequences. The scalar $k$ balances the two terms. Empirically, this objective correlates strongly with lower heart rate error across datasets. The search runs over a small grid of $(\lambda, \alpha)$ values and requires no training data. Once selected, $(\lambda^*, \alpha^*)$ are held fixed and used to compute final heart-rate estimates. Additional implementation details are provided in \autoref{sec:prism}.

\subsection{Datasets}
To assess the performance of PRISM under realistic conditions, including motion and varied illumination, we utilize the two standard rPPG benchmark datasets.

\textbf{PURE}~\cite{pure}: 
A widely used real-world dataset consisting of RGB videos collected using eco274CVGE camera at resolution ($640\times480$, 30\,Hz) from 10 subjects (8 male, 2 female) performing six head motion tasks ranging from steady sitting to talking, translation, and rotation. Synchronized ground truth PPG and SpO$_{2}$ signals were recorded using a finger-clip oximeter at 60\,Hz .

\textbf{UBFC-rPPG}~\cite{ubfc-rppg}:
A commonly used dataset consisting of 42 RGB videos ($640\times480$, 30\,Hz) recorded indoors with a Logitech C920 HD Pro webcam under natural and artificial lighting. PPG signals were simultaneously captured using a finger-clip pulse oximeter.

\subsection{Implementation Enhancements to rPPG-Toolbox}
\label{sec:toolbox_modifications}

We used the rPPG-toolbox~\cite{rppg-toolbox} to ensure reproducibility, but implemented several modifications to enhance heart rate estimation accuracy across all methods. 

\textbf{Frequency Resolution:} We increased the FFT bin count from $N = 2^9 = 512$ bins (frequency resolution of approximately 3.5~bpm) to $N = 2^{14}$, achieving a substantially finer resolution of approximately 0.1~bpm. 

\textbf{Temporal Windowing:} We replaced the default 60-second analysis windows with 10-second non-overlapping windows. Shorter windows better track natural heart rate variability and provide more temporally precise measurements, particularly during motion scenarios.

\textbf{Face Detection:} Our ablation studies revealed that YOLOv5~\cite{yolov5}, a learning-based face detector already available in the toolbox, significantly outperforms the default HaarCascade method. Superior detection accuracy ensures consistent face region-of-interest (ROI) extraction, reducing signal contamination from background pixels.

\textbf{Dynamic Face Tracking:} We enabled dynamic face tracking to continuously adapt to subject movement throughout the video duration, replacing the toolbox's default static tracking assumption. Continuous adaptation to head motion maintains ROI alignment, which is critical for motion-heavy scenarios like PURE.

\textbf{ROI Configuration:} We configured the bounding box parameter by setting \texttt{largebox = False}, thereby minimizing background interference and ensuring more accurate extraction of the facial region of interest.

These modifications improve the performance of all evaluated methods, as demonstrated in \autoref{sec:ablation}.

\section{Results and Discussion}
\label{sec:results}

\begin{figure}
\centering
\includegraphics[width=1.0\columnwidth]{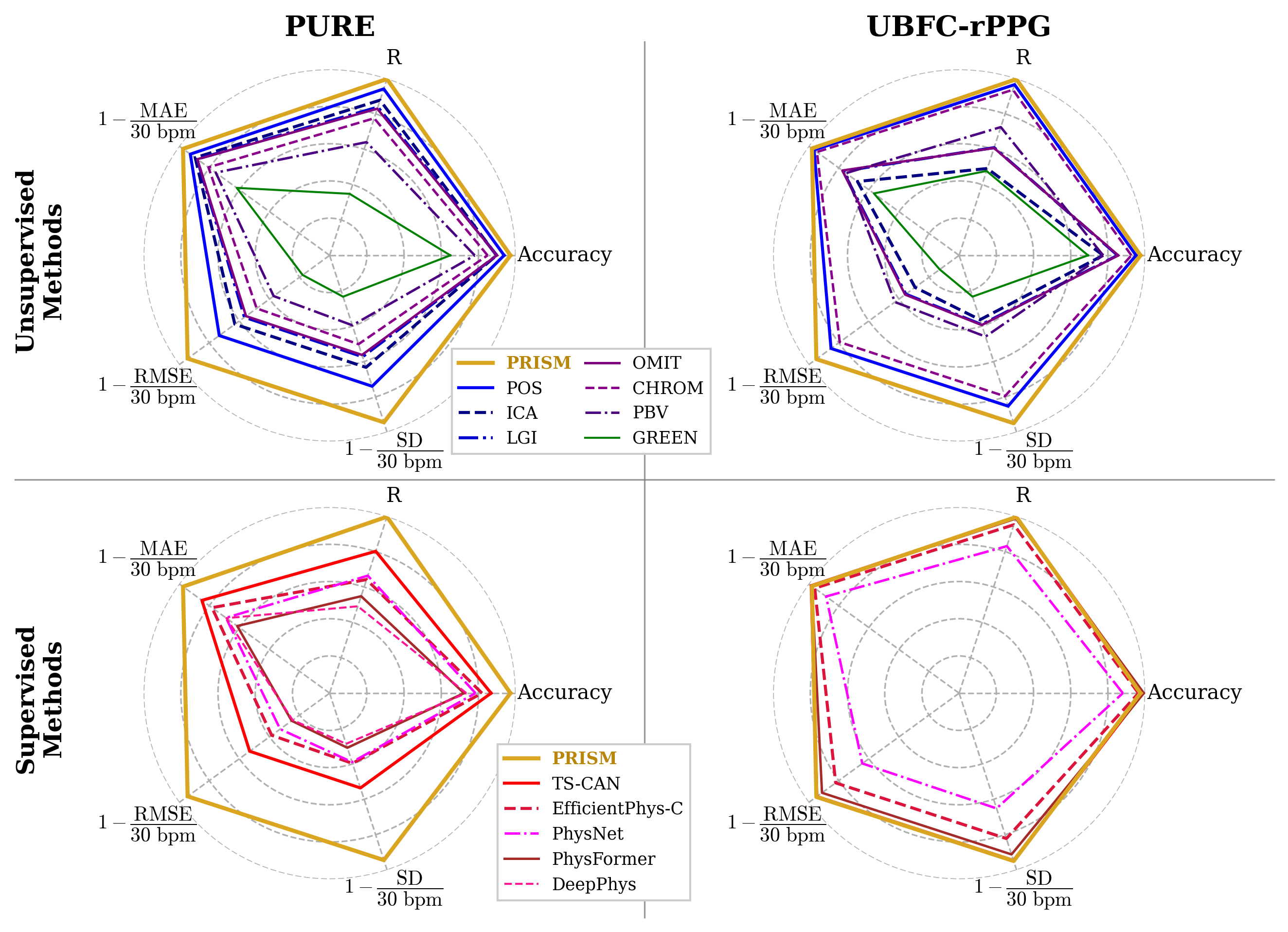}
\caption{Radar plots comparing performance across five evaluation metrics on the PURE and UBFC-rPPG datasets. PRISM outperforms other methods across most of the metrics. All metrics are normalized such that larger area indicates better performance.}
\label{fig:spider}
\vspace*{-10pt}
\end{figure}

We evaluate performance using five metrics: Mean Absolute Error (MAE), Root Mean Square Error (RMSE), standard deviation (SD), Pearson correlation (R), and accuracy within a 5 bpm threshold, where predictions within $\pm 5$ bpm of the ground truth are considered correct. This method of computing the accuracy provides an intuitive performance metric that aligns with clinically acceptable error ranges for heart rate monitoring and is more readily interpretable by medical practitioners than traditional correlation-based metrics. They are computed on non-overlapping 10-second windows to reflect real-time tracking performance and account for natural heart rate variability, results are shown in \autoref{tab:all_metrics}. \autoref{fig:spider} summarizes performance across all methods using radar plots, with each axis normalized such that larger area indicates better performance.

PRISM demonstrates superior performance among other unsupervised methods across both datasets. On PURE, PRISM achieves the lowest MAE, 0.77 bpm, and RMSE, 1.69 bpm, significantly outperforming the previous best unsupervised method, POS, which achieved 2.23 bpm MAE and 8.00 RMSE. This corresponds to a 65\% reduction in MAE and 79\% reduction in RMSE. On UBFC-rPPG, PRISM achieves 0.66 bpm MAE and 1.54 bpm RMSE, again outperforming POS's 1.08 bpm MAE while improving RMSE by a substantial 65\%. 

The accuracy results further demonstrate PRISM's effectiveness, achieving 97.3\% accuracy on PURE and 97.5\% accuracy on UBFC-rPPG, outperforming all other unsupervised methods. Notably, PRISM's performance approaches that of state-of-the-art supervised methods while maintaining the computational efficiency and deployment advantages of unsupervised approaches. Namely, on UBFC-rPPG, PRISM matches the accuracy of 97.5\% of DeepPhys and approaches TSCAN's 0.51 bpm MAE performance.

Statistical analysis using paired t-tests confirms PRISM's performance on UBFC-rPPG is statistically equivalent to leading supervised methods, TS-CAN ($p = 0.208$), PhysFormer ($p = 0.880$), and DeepPhys ($p = 0.595$). 
Thus, PRISM achieves statistically comparable performance to deep learning approaches without requiring training data.

Both PURE and UBFC-rPPG datasets contain natural varying lighting conditions, demonstrating PRISM's adaptability compared to methods designed for artificial illumination or via invariance priors. Additionally, \autoref{sec:pure_per_condition} presents PRISM's per-condition results on PURE, showing robust performance across different motion scenarios.

These results validate PRISM's adaptive parameter selection strategy, demonstrating that online optimization of color mixing and temporal filtering parameters can significantly improve rPPG extraction quality across diverse experimental conditions and datasets. 

\begin{table}[H]
\centering
\caption{Comprehensive performance comparison across all metrics on the PURE and UBFC-rPPG benchmark datasets. PRISM's performance is highlighted. MAE, RMSE, and SD are in bpm. R denotes Pearson correlation. Acc. is the percentage of estimates within a $\pm5~\text{bpm}$ threshold.}
\label{tab:all_metrics}
\renewcommand{\arraystretch}{0.9}
\setlength{\tabcolsep}{4pt}
\footnotesize
\begin{adjustbox}{max width=\textwidth}
\begin{tabular}{lrcccccccccc}
\toprule
 & & \multicolumn{10}{c}{\textbf{Test Set}} \\
 & & \multicolumn{5}{c}{PURE~\cite{pure}} & \multicolumn{5}{c}{UBFC-rPPG~\cite{ubfc-rppg}} \\
\cmidrule(lr){3-7} \cmidrule(lr){8-12} 
 & \textbf{Method} & MAE$^\downarrow$ & RMSE$^\downarrow$ & SD$^\downarrow$ & R$^\uparrow$ & Acc.$^\uparrow$ & MAE$^\downarrow$ & RMSE$^\downarrow$ & SD$^\downarrow$ & R$^\uparrow$ & Acc.$^\uparrow$ \\  
\midrule
\multirow{8}{*}{\rotatebox[origin=c]{90}{Unsupervised}}
& GREEN \cite{green} & 11.48 & 24.62 & 22.99 & 0.348 & 64.9\% & 13.01 & 26.13 & 23.00 & 0.477 & 69.3\% \\
& ICA \cite{ica} & 3.20 & 11.11 & 11.01 & 0.877 & 89.7\% & 9.73 & 21.21 & 19.09 & 0.491 & 77.4\% \\
& CHROM \cite{chrom} & 5.84 & 15.45 & 14.94 & 0.775 & 84.9\% & 1.65 & 6.20 & 6.11 & 0.936 & 92.6\% \\
& PBV \cite{pbv} & 7.30 & 18.84 & 18.14 & 0.640 & 78.0\% & 7.44 & 17.20 & 16.23 & 0.725 & 77.4\% \\
& OMIT \cite{omit} & 3.62 & 13.33 & 13.12 & 0.829 & 89.9\% & 6.72 & 19.24 & 18.19 & 0.607 & 85.5\% \\
& POS \cite{pos} & 2.23 & 8.00 & 7.81 & 0.941 & 93.8\% & 1.08 & 4.46 & 4.44 & 0.965 & 95.1\% \\
& LGI \cite{lgi} & 3.58 & 13.02 & 12.81 & 0.838 & 89.9\% & 6.88 & 19.37 & 18.28 & 0.609 & 85.2\% \\
& \databg{\textbf{PRISM (Ours)}} & \databg{\textbf{0.77}} & \databg{\textbf{1.69}} & \databg{\textbf{1.69}} & \databg{\textbf{0.997}} & \databg{\textbf{97.2\%}} & \databg{0.66} & \databg{\textbf{1.54}} & \databg{\textbf{1.54}} & \databg{\textbf{0.996}} & \databg{97.5\%} \\
\midrule
\multirow{5}{*}{\rotatebox[origin=c]{90}{Supervised}}
& DeepPhys \cite{deepphys} & 9.42 & 22.62 & 21.41 & 0.491 & 73.5\% & 0.58 & 1.70 & 1.69 & 0.995 & 97.5\% \\
& PhysNet \cite{physnet} & 9.33 & 20.13 & 18.31 & 0.664 & 78.4\% & 3.48 & 10.71 & 10.46 & 0.831 & 87.9\% \\
& TS-CAN \cite{ts-can} & 4.54 & 14.06 & 13.93 & 0.802 & 86.8\% & \textbf{0.51} & 1.64 & 1.64 & 0.995 & 98.6\% \\
& PhysFormer \cite{physformer} & 11.58 & 22.43 & 20.74 & 0.548 & 72.2\% & 0.64 & 2.66 & 2.66 & 0.987 & \textbf{99.3\%} \\
& EfficientPhys-C \cite{efficientphys} & 6.55 & 18.45 & 18.10 & 0.642 & 81.9\% & 1.22 & 5.39 & 5.35 & 0.952 & 96.4\% \\
\bottomrule
\end{tabular}
\end{adjustbox}
\end{table}

\section{Conclusions}
PRISM demonstrates that unsupervised methods can match or exceed supervised performance through adaptive optimization. While these results are promising, the current evaluation focuses primarily on lighter skin tones present in the tested datasets, and more comprehensive testing across diverse skin tones is needed to fully validate the method's generalizability. Additionally, ongoing work is testing PRISM in more challenging outdoor and robot-mounted scenarios where environmental conditions are completely uncontrolled, which will provide further insights into the method's robustness and outdoor deployment potential. We also plan to integrate PRISM in the rPPG-Toolbox repository for easier access and reproducibility.

\begin{ack}
This work has been partially supported by the Defense Advance Research Projects Agency (award HR00112420329) and National Science Foundation (awards 2427948 and 2406231). 

\end{ack}

{\small
    \bibliographystyle{unsrtnat}
    \bibliography{main}
}

\appendix

\begin{table}[H]
\label{table:ablation_fixed_lambda}
\end{table}

\section{Ablation}
\label{sec:ablation}    
To understand PRISM's performance and validate our design choices,  we performed an extensive ablation study, presented in \autoref{tab:ablation_comprehensive}. We evaluated several variants of our method:
\begin{itemize}
    \item \textbf{Full PRISM:} The complete model with joint optimization of both the color mixing weight $\alpha$ and the smoothness parameter $\lambda$.
    \item \textbf{Best Fixed $\alpha$ / $\lambda$:} The model with one parameter fixed, while the other is optimized. This isolates the contribution of adapting each parameter individually.
    \item \textbf{Concentration-only:} The model optimizes the objective function without the temporal variation penalty ($k=0$), relying solely on spectral concentration.
    \item \textbf{TV-only:} Model optimizes the objective function without concentration value.
\end{itemize}

The results demonstrate that each component of PRISM is critical. The full model consistently achieves the lowest MAE, indicating the most reliable performance. While fixing $\alpha$ or $\lambda $ can sometimes give a lower RMSE by avoiding occasional large errors coming from a subject in the dataset, the higher MAE shows a decrease in accuracy. A crucial observation, the optimal fixed parameters differ significantly between datasets (best fixed $\alpha$ is 0.6 for PURE but 0.8 for UBFC-rPPG) this verifies the need for an adaptive approach to generalize across different conditions.

Removing the temporal variation penalty shoots up the error rates, confirming the need for penalty in our objective function. Removing the concentration had the most significant impact, confirming its critical role in guiding the optimization toward physiologically relevant frequencies. We hypothesize this is because the concentration term $C$ acts as a frequency-domain constraint that distinguishes genuine pulse signals from noise and artifacts. Without this physiological prior, the temporal variation penalty alone cannot differentiate between smooth but non-physiological signals, such as low-frequency illumination changes, and actual cardiac activity, leading to parameter selections that produce temporally consistent but physiologically meaningless results.

\begin{table}[H]
\centering
\caption{Comprehensive ablation study of PRISM's components on the PURE and UBFC-rPPG datasets. We compare the full model against variants with fixed parameters or a modified objective function. The `Best Fixed' settings show the result for the single best-performing fixed parameter value from our experiments. `Concentration-only' refers to optimizing without the temporal variation penalty ($k=0$). `TV-Only' uses optimization function without $C$.}
\label{tab:ablation_comprehensive}
\begin{adjustbox}{max width=\textwidth}
\begin{tabular}{llccccr}
\toprule
\textbf{Dataset} & \textbf{Configuration} & \textbf{MAE$^\downarrow$} & \textbf{RMSE$^\downarrow$} & \textbf{Acc.$^\uparrow$} & \textbf{Selected $\boldsymbol{\lambda}$} & \textbf{Selected $\boldsymbol{\alpha}$} \\
\midrule
\multirow{5}{*}{\rotatebox[origin=c]{90}{PURE}} 
& \databg{Full PRISM (Ours)} & \databg{\textbf{0.77}} & \databg{1.69} & \databg{97.3\%} & \databg{Adaptive} & \databg{Adaptive} \\
\cmidrule{2-7}
& Best Fixed $\alpha$ ($\alpha=0.6$) & 0.82 & \textbf{1.26} & \textbf{97.5\%} & 0.155 & 0.6  \\
& Best Fixed $\lambda$ ($\lambda=0.5$) & 0.99 & 1.74 & 96.4\% & 0.5 & 0.466 \\
& Concentration-only ($k=0$) & 4.02 & 5.47 & 85.8\% & 0.107 & 0.171 \\
& TV-only ($C=0$) & 14.30 & 16.55 & 64.2\% & 0.01  & 0.0 \\
\midrule
\multirow{5}{*}{\rotatebox[origin=c]{90}{UBFC-rPPG}}
& \databg{Full PRISM (Ours)} & \databg{\textbf{0.66}} & \databg{1.54} & \databg{97.5\%} & \databg{Adaptive} & \databg{Adaptive} \\
\cmidrule{2-7}
& Best Fixed $\alpha$ ($\alpha=0.8$) & 0.70 & \textbf{1.15} & \textbf{97.6\%} & 0.058 & 0.8  \\
& Best Fixed $\lambda$ ($\lambda=0.05$) & 1.04 & 1.94 & 96.6\% & 0.05 & 0.560 \\
& Concentration-only ($k=0$) & 1.83 & 2.93 & 93.6\% & 0.052 & 0.479 \\
& TV-only ($C=0$) & 4.74 & 8.33 & 81.9\% & 0.01  & 0.0  \\
\bottomrule
\end{tabular}
\end{adjustbox}
\end{table}

\section{PRISM implementation details}
\label{sec:prism}

The PRISM algorithm is built as a drop-in replacement for traditional projection-based rPPG pipelines such as POS, but with online parameter adaptation for robustness under varying illumination and motion. It begins with preprocessing and normalization, followed by an online grid-search of the optimal color projection and heart rate estimation.

Each input video is first processed with face detection \cite{yolov5}. In each frame, a rectangle region of interest is defined over the face, and pixel intensities within this region are averaged for each RGB channel, producing the raw signals $R(t)$, $G(t)$, and $B(t)$. To suppress illumination drift while preserving pulsatile components, each channel is normalized by dividing it by a smooth estimate of its baseline. For each $X(t) \in \{R(t), G(t), B(t)\}$, we compute
\begin{equation}
    \hat X(t) = \frac{X(t)}{\tilde X(t)},
\end{equation}
where $\tilde X(t)$ is a smoothing spline fit to $X(t)$ by minimizing the residual functional \cite{spline_original,spline_theory},
\begin{equation}
    \tilde X = \arg\min_{h \in \mathcal{H}^2} \left\{ \sum_{k=1}^K w_k \left(h(t_k) - X(t_k)\right)^2 + \lambda \int \left(h''(t)\right)^2 dt \right\} \ .
\end{equation}
Here, $\mathcal{H}^2$ denotes the Sobolev space of twice-differentiable functions, $w_k$ are frame weights (set uniformly), and $\lambda > 0$ controls the smoothness. Smaller values of $\lambda$ allow the spline to follow fluctuations more closely, while larger values enforce a smoother baseline.

Once normalized, we construct a pulse signal $s(t)$ as a linear combination of the normalized color channels. In the Plane-Orthogonal-to-Skin (POS) method, the signal is given by
\begin{equation}
    s_{\text{POS}}(t) = s_1(t) + \gamma \cdot s_2(t) \ ,
\end{equation}
where $s_1(t) = \hat G(t) - \hat B(t)$, $s_2(t) = \hat G(t) + \hat B(t) - 2\hat R(t)$, and $\gamma$ is adaptively set as the ratio of standard deviations $\sigma(s_1)/\sigma(s_2)$. Expanding and regrouping terms, we obtain
\begin{equation}
    s_{\text{POS}}(t) = (\gamma + 1)\hat G(t) + (\gamma - 1)\hat B(t) - 2\gamma \hat R(t) \ .
\end{equation}
We can factor out $(1 + \gamma)$ and introduce $\alpha = (1 - \gamma) / (1 + \gamma)$ to obtain an equivalent form:
\begin{equation}
    s_{\text{PRISM}}(t) = \hat G(t) - \left( \alpha \cdot \hat B(t) + (1 - \alpha) \cdot \hat R(t) \right) \ ,
\end{equation}
with $\alpha \in [-1, 1]$ to be selected adaptively to optimize for signal quality. The formulation retains interpretability while enabling data-driven projection that adapts to varying spectral and motion conditions.

To effectively extract the pulse signal across varying conditions, the parameters $\alpha$ and $\lambda$ must be selected to balance noise suppression, spectral fidelity, and robustness to motion and illumination changes. Rather than relying on fixed values, PRISM adaptively tunes these parameters based on the observed signal characteristics. This is achieved by segmenting the input stream into consecutive non-overlapping windows of duration $\delta t = 10~\text{s}$, over which signal quality can be locally assessed.

Within each window, the normalized signal $s(t)$ is computed for candidate $(\lambda, \alpha)$ pairs. A fast Fourier transform (FFT) is performed to extract the dominant frequency to give an estimated heart rate $h_i$. These estimates form the basis for an objective function that encourages sharp spectral peaks in the physiological band and smooth heart rate evolution over time.

To select the optimal parameters $(\lambda^*, \alpha^*)$, PRISM defines an objective that balances two criteria: spectral concentration of the pulse signal and temporal stability of the estimated heart rates. The spectral concentration $C$ is defined as the fraction of signal power falling within some physiological band $[f_\text{min}, f_\text{max}]$:
\begin{equation}
    C = \frac{\int_{f_\text{min}}^{f_\text{max}} |S(f)|^2 \, df}{\int_0^{f_\text{nyquist}} |S(f)|^2 \, df} \ ,
\end{equation}
where $S(f)$ is the Fourier transform of $s(t)$ and $f_\text{nyquist} = \frac{1}{2}f_\text{frame}$ is the Nyquist frequency, equal to half the frame rate. Temporal variation is computed as
\begin{equation}
    \mathrm{TV} = \frac{1}{t_n - t_1} \sum_{i=1}^{n - 1} | h_{i+1} - h_i | \ ,
\end{equation}
where $t_1$ and $t_n$ are the midpoints of the first and last windows. The joint objective minimized by PRISM is
\begin{equation}
    \mathcal{L}(\lambda, \alpha) = k \cdot \mathrm{TV}(\lambda, \alpha) - C(\lambda, \alpha) \ ,
\end{equation}
with $k = 1/3$, set empirically as a heuristic. This objective function encourages signals with high spectral concentration in the plausible physiological range and smooth trajectories of estimated heart rate over time. In practice, a discrete grid search is performed over $\alpha \in \{0.5, 0.6, 0.7, 0.8, 0.9, 1.0\}$ and $\lambda \in \{0.01, 0.05, 0.1, 0.5, 1.0\}$. The parameter pair minimizing $\mathcal{L}$ is selected and used to compute final estimates. Empirically, $\alpha$ values below 0.5 are seldom optimal and are therefore excluded from the search. This is intuitive as a large $\alpha$ corresponds to the primary projection given in POS (i.e., $s_1(t) = \hat G(t) - \hat B(t)$), meanwhile a small $\alpha$ corresponds closer to the secondary projection (i.e., $s_2(t) = \hat G(t) +\hat B(t)- 2\hat R(t)$). 

Occasionally, harmonic frequencies dominate the power spectra of rPPG signals, which can result in predictions that are twice the true heart rate frequency. To alleviate this, two bands are used for the spectral concentration \(C\): a high band \([0.75\,\mathrm{Hz},\,4.0\,\mathrm{Hz}]\) and a low band \([0.5\,\mathrm{Hz},\,3.0\,\mathrm{Hz}]\).
An rPPG candidate is derived from each band, \(s_{\text{PRISM},H}(t)\) and \(s_{\text{PRISM},L}(t)\), and the estimated windowed heart-rate sequences are given as: \(\{f_H^{(k)}\}_{k=1}^K\) and \(\{f_L^{(k)}\}_{k=1}^K\).
Let
\begin{equation}
\label{eq:band-means}
\bar f_H \triangleq \frac{1}{K}\sum_{k=1}^K f_H^{(k)}, \qquad
\bar f_L \triangleq \frac{1}{K}\sum_{k=1}^K f_L^{(k)}.
\end{equation}
Denote the high-band estimate as a harmonic of the low-band estimate if $\bar f_H \approx 2\,\bar f_L$. The final selection is
\begin{equation}
    \label{eq:final-selection}
    \widehat{s}_{\text{PRISM}}(t) =
    \begin{cases}
    s_{\text{PRISM},L}(t), & \text{if } \bar f_H \approx 2\,\bar f_L,\\[4pt]
    s_{\text{PRISM},H}(t), & \text{otherwise.}
    \end{cases}
\end{equation}

Since parameter selection relies only on recent signal windows, PRISM operates in an online manner. After a brief initialization period of roughly one minute, the system converges to stable parameter values and can thereafter update them periodically as conditions change, or maintain them if the environment remains stable. Moreover, inference can begin immediately with predefined $\alpha$ and $\lambda$ values, which, as shown in \autoref{sec:ablation}, already provide competitive performance. Subsequent adaptation then refines these parameters to better match the observed data.

\section{Per-condition performance on PURE}
\label{sec:pure_per_condition}

\begin{table}[H]
\centering
\caption{Per-condition performance of PRISM on the PURE dataset. MAE, RMSE, and SD are reported in beats per minute (bpm). R is the Pearson correlation coefficient. Acc. is the percentage of estimates within a $\pm5~\text{bpm}$ threshold.}
\label{tab:pure_condition_averages}
\begin{tabular}{lrrrrr}
\toprule
{} &  MAE$^\downarrow$ &  RMSE$^\downarrow$ &  SD$^\downarrow$ & R$^\uparrow$ & Acc.$^\uparrow$ \\
Condition            &           &            &            &            &        \\
\midrule
Steady               & 0.48 & 0.72 & 0.72 & 0.999 & 100.00  \\
Talking              & 1.48 & 3.00 & 2.93 & 0.993 &  88.89 \\
Slow Translation     & 0.52 & 0.78 & 0.77 & 0.999 & 100.00 \\
Fast Translation     & 1.26 & 2.50 & 2.50 & 0.993 &  93.75  \\
Small Head Rotation  & 0.40 & 0.54 & 0.54 & 1.000 & 100.00  \\
Medium Head Rotation & 0.53 & 0.95 & 0.95 & 0.999 & 100.00 \\
\bottomrule
\end{tabular}
\end{table}

\section{Ablation: Toolbox Modifications vs. Algorithm Contribution}
\label{sec:ablation}

To isolate PRISM's algorithmic contribution from implementation improvements, we evaluated our method under two configurations:

\textbf{Standard rPPG-Toolbox Configuration:}
Using default settings (512 FFT bins, 60-second windows, HaarCascade detection, static tracking), PRISM achieves:
\begin{itemize}
    \item PURE: MAE = 2.66~bpm
    \item UBFC-rPPG: MAE = 3.10~bpm
\end{itemize}

\textbf{Enhanced Configuration (reported in \autoref{tab:all_metrics}):}
With our modifications (\autoref{sec:toolbox_modifications}), PRISM achieves:
\begin{itemize}
    \item PURE: MAE = 0.77~bpm (71\% improvement)
    \item UBFC-rPPG: MAE = 0.66~bpm (79\% improvement)
\end{itemize}

Importantly, our modifications improve the performance of unsupervised methods while showing no statistical difference on supervised approaches. \autoref{tab:ablation} demonstrates the impact across representative methods. For unsupervised methods like POS, the enhanced configuration provides consistent improvements on both datasets (39\% improvement on PURE, 75\% on UBFC-rPPG). Interestingly, the supervised method TS-CAN shows degraded performance on PURE with the enhanced configuration (23\% worse), while improving substantially on UBFC-rPPG (60\% improvement). This suggests that supervised methods may have been optimized for the specific characteristics of the standard toolbox settings, whereas unsupervised methods benefit more uniformly from improved signal preprocessing. PRISM's adaptive optimization demonstrates robust performance under both configurations, achieving competitive results with standard settings and state-of-the-art performance with enhanced settings.

\begin{table}[h]
\centering
\caption{Performance comparison of leading unsupervised and supervised methods under standard versus enhanced rPPG-Toolbox configurations. Unsupervised methods (US) benefit consistently from enhanced settings, while supervised methods (S) show dataset-dependent responses, suggesting possible overfitting to standard toolbox characteristics.}
\label{tab:ablation}
\footnotesize
\begin{tabular}{llcccc}
\toprule
 & & \multicolumn{2}{c}{\textbf{PURE}} & \multicolumn{2}{c}{\textbf{UBFC-rPPG}} \\
\cmidrule(lr){3-4} \cmidrule(lr){5-6}
& \textbf{Method} & \textbf{Standard} & \textbf{Enhanced} & \textbf{Standard} & \textbf{Enhanced} \\
& & MAE (bpm) & MAE (bpm) & MAE (bpm) & MAE (bpm) \\
\midrule
\multirow{2}{*}{\rotatebox[origin=c]{90}{\footnotesize US}}
& POS \cite{pos} & 3.67 & 2.23 & 4.00 & 1.08 \\
& \textbf{PRISM} & \textbf{2.66} & \textbf{0.77} & \textbf{3.10} & \textbf{0.66} \\
\midrule
\multirow{1}{*}{\rotatebox[origin=c]{90}{\footnotesize S}}
& TS-CAN \cite{ts-can} & 3.69 & 4.54 & 1.29 & 0.51 \\
\bottomrule
\end{tabular}
\end{table}





\end{document}